\documentclass[runningheads]{llncs}
\usepackage{enumitem}
\usepackage{times}
\usepackage{epsfig}
\usepackage{graphicx}
\usepackage{amsmath}
\usepackage{amssymb}
\usepackage{tabularx}
\usepackage{multirow}
\usepackage{multicol}
\usepackage{array, threeparttable, caption, booktabs}
\usepackage{pifont}
\usepackage{float}
\usepackage{subfigure}
\usepackage{placeins}
\usepackage{pifont}
\usepackage{cite}

\newcommand{\cmark}{\ding{51}}%
\newcommand{\xmark}{\ding{55}}%
\newcommand{\iSegFormer}{\texttt{iSegFormer}}

\usepackage[pagebackref=true,breaklinks=true,letterpaper=true,colorlinks,bookmarks=false]{hyperref}

\begin{document}

\title{iSegFormer: Interactive Segmentation via Transformers with Application to 3D Knee MR Images}

\author{Qin Liu, Zhenlin Xu, Yining Jiao, Marc Niethammer}
\institute{Department of Computer Science \\ University of North Carolina at Chapel Hill \\ \email{\{qinliu19,zhenlinx,jyn,mn\}@cs.unc.edu}}

\maketitle

\begin{abstract}
Interactive image segmentation has been widely applied to obtain high-quality voxel-level labels for medical images.
The recent success of Transformers on various vision tasks has paved the road for developing Transformer-based interactive image segmentation approaches.
However, these approaches remain unexplored and, in particular, have not been developed for 3D medical image segmentation. To fill this research gap, we investigate Transformer-based interactive image segmentation and its application to 3D medical images.
This is a nontrivial task due to two main challenges: 1) limited memory for computationally inefficient Transformers and 2) limited labels for 3D medical images.
To tackle the first challenge, we propose {\iSegFormer}, a memory-efficient Transformer that combines a Swin Transformer with a lightweight multilayer perceptron (MLP) decoder.
To address the second challenge, we pretrain {\iSegFormer} on large amount of unlabeled datasets and then finetune it with only a limited number of segmented 2D slices.
We further propagate the 2D segmentations obtained by {\iSegFormer} to unsegmented slices in 3D images using a pre-existing segmentation propagation model pretrained on videos. We evaluate {\iSegFormer} on the public OAI-ZIB dataset for interactive knee cartilage segmentation. Evaluation results show that {\iSegFormer} outperforms its convolutional neural network (CNN) counterparts on interactive 2D knee cartilage segmentation, with competitive computational efficiency. When propagating the 2D interactive segmentations 
of 5 slices to other unprocessed slices within the same 3D volume, we achieve 82.2\% Dice score for 3D knee cartilage segmentation. Code is available at
\href{https://github.com/uncbiag/iSegFormer}{https://github.com/uncbiag/iSegFormer}.
\end{abstract}

\section{Introduction}

Deep learning-based approaches for semantic and instance segmentation have achieved tremendous success for medical images~\cite{wang2022medical,shen2017deep}. 
However, these approaches are data hungry and heavily rely on the availability of large-scale voxel-level segmentations, which to obtain requires significant labor, time, and expertise~\cite{tajbakhsh2020embracing}. 
Interactive image segmentation, which aims to extract image objects using limited human interactions, is an efficient way to obtain these segmentations~\cite{xu2016deep}. 
Hence, significant work is ongoing to explore interactive segmentation approaches~\cite{xu2016deep,xu2017deep,sofiiuk2021reviving}.

Existing state-of-the-art interactive segmentation methods are all CNN-based, leveraging the good representation ability of CNNs~\cite{sofiiuk2021reviving,zhang2020interactive}.
Although these CNN-based methods have achieved excellent performance, they suffer from limited receptive fields and cannot learn global and long-range semantic information well due to the inductive bias of locality and weight sharing~\cite{cohen2016inductive}.
Several techniques have been proposed to address these problems, such as atrous convolutional layers~\cite{chen2017deeplab} and non-local blocks~\cite{wang2018nonlocal}.
A recent research direction is to replace CNN with vision Transformer (ViT), which can naturally capture long-range dependencies through the self-attention mechanism~\cite{yuan2021hrformer,xie2021segformer}.
Following this direction, various vision Transformers have been proposed for medical image segmentation~\cite{cao2021swin}, paving the road for developing Transformer-based interactive image segmentation approaches. However, these approaches remain unexplored and, in particular, have not been developed for 3D medical images.

In this work, we aim to fill this research gap by investigating Transformer-based interactive image segmentation and its application to 3D medical images.
This is a challenging task due to: 1) limited memory for computationally inefficient Transformers and 2) limited labels for 3D medical images.
To tackle the first challenge, we propose {\iSegFormer}, a memory-efficient Transformer that combines a Swin Transformer with a lightweight multilayer perceptron (MLP) decoder. With the efficient Swin Transformer blocks for hierarchical self-attention and the simple MLP decoder for aggregating both local and global attention, {\iSegFormer} learns powerful representations while achieving high computational efficiencies.
To address the second challenge, we pretrain {\iSegFormer} on large amount of unlabeled datasets and then finetune it with only a limited number of segmented 2D slices.
To extend {\iSegFormer} to 3D interactive segmentation, we further combine it with a segmentation propagation module that propagates segmented 2D slices to unlabeled ones in the same image volume.
When the propagated segmentations are not as desired, the user can refine them and start a new round of propagation if necessary.
Specifically, we combine {\iSegFormer} with STCN~\cite{cheng2021rethinking}, which achieves state-of-the-art results on interactive video object segmentation. We use a pretrained STCN model without finetuning on medical images.

We evaluate {\iSegFormer} on the public OAI-ZIB dataset for interactive knee cartilage segmentation. Evaluation results show that {\iSegFormer} outperforms its convolutional neural network (CNN) counterparts on interactive 2D knee cartilage segmentation, with competitive computational efficiency. When propagating 2D interactive segmentation of 5 slices to other unprocessed slices within the same 3D volume, the propagation model achieves a Dice score of 82.2\% for 3D knee cartilage segmentation.
Finally, we show that {\iSegFormer} combined with the segmentation propagation model results in an efficient framework for interactive 3D medical image segmentation.

Our contributions are as follows:
\begin{itemize}
  \item[1)] We propose {\iSegFormer}, a memory-efficient Transformer that combines a Swin Transformer with a lightweight MLP decoder, for interactive image segmentation. 
  \item[2)] {\iSegFormer} outperforms its CNN counterparts for interactive 2D knee cartilage segmentation on the OAI-ZIB dataset with comparable computational efficiency with CNNs. To the best of our knowledge, {\iSegFormer} is the first Transformer-based approach for interactive medical image segmentation.
  \item[3)] We further show that {\iSegFormer} can be easily extended to interactive 3D medical image segmentation by combining it with a pre-existing segmentation propagation model trained on videos.
\end{itemize}

\section{Related Work}
\label{sec:related_work}

\textbf{Interactive Medical Image Segmentation}
Existing interactive segmentation methods for medical images are all CNN-based~\cite{diaz2022monai,chao2020interactive,sakinis2019interactive,luo2021mideepseg}, partially inspired by the seminal work~\cite{xu2016deep}.
MIDeepSeg~\cite{luo2021mideepseg} proposes a click-based approach that encodes foreground and background clicks through Gaussian-smoothed maps, which serve as the input to the CNN encoder-decoder. 
Recently, MONAI Label~\cite{diaz2022monai} proposes an open-source framework for CNN-based interactive segmentation of 3D medical images, which consists of both click-based and scribbles-based interactive segmentation algorithms.
In this work, we are interested in Transformer-based interactive segmentation, which has not been well-explored, especially for medical images.
Given the recent success of vision Transformers for automatic medical image segmentation~\cite{chen2021transunet}, it is a natural extension for applying Transformers to interactive medical image segmentation.
Specifically, we apply to the challenging knee cartilage segmentation from MR images~\cite{liu2019multi}.

\noindent\textbf{Vision Transformers}
The vision Transformer (ViT)~\cite{dosovitskiy2020image} first shows that a pure Transformer can achieve state-of-the-art performance for image classification. 
Pyramid vision Transformer~\cite{wang2021pyramid} further shows that ViT can also achieve comparable performance with CNNs in dense prediction tasks. 
SegFormer~\cite{xie2021segformer} proposes an efficient segmentation approach that combines a hierarchically structured Transformer encoder with a light-weight decoder using MLPs, demonstrating the state-of-the-art segmentation performance compared with CNNs.
The Swin Transformer~\cite{liu2021swin} is a breakthrough that shows the superiority of hierarchical vision Transformer over CNN as a general vision backbone.
Meanwhile, different ViTs have also been proposed in automatic medical image segmentation~\cite{cao2021swin,zhang2021transfuse,gao2021utnet}.
Among these methods, Swin-Unet~\cite{cao2021swin} and UTNet~\cite{gao2021utnet} are ``U-shaped" networks inspired by Unets~\cite{ronneberger2015u}.
Comparing with these methods, {\iSegFormer} is more efficient considering its efficient Swin Transformer encoder and light-wighted MLP decoder.

\noindent\textbf{Interactive Video Object Segmentation (iVOS)} 
iVOS aims at extracting high-quality segmentation masks of a target video object through two modules: a 2D interactive segmentation module and a segmentation propagation module~\cite{oh2019fast,cheng2021modular}.
MiVOS~\cite{cheng2021modular} decouples the two modules and train them independently. 
During inference, MiVOS first interactively segments one or several frames in a video, followed by propagating the segmented frames to the unsegmented ones.
STCN~\cite{cheng2021rethinking} further improves the segmentation propagation module in MiVOS by directly encoding the query and memory frames without re-encoding the mask features for every object.
Inspired by the observation that iVOS pipeline can be directly applied to 3D medical images, we extend {\iSegFormer} to interactive 3D medical image segmentation by combining it with an existing STCN model trained on videos, leading to a very promissing results even without fine-tuning.


\section{Method}
\label{sec:method}

\begin{figure}[ht]
  \centering
  \includegraphics[width=12.0cm, height=7.0cm]{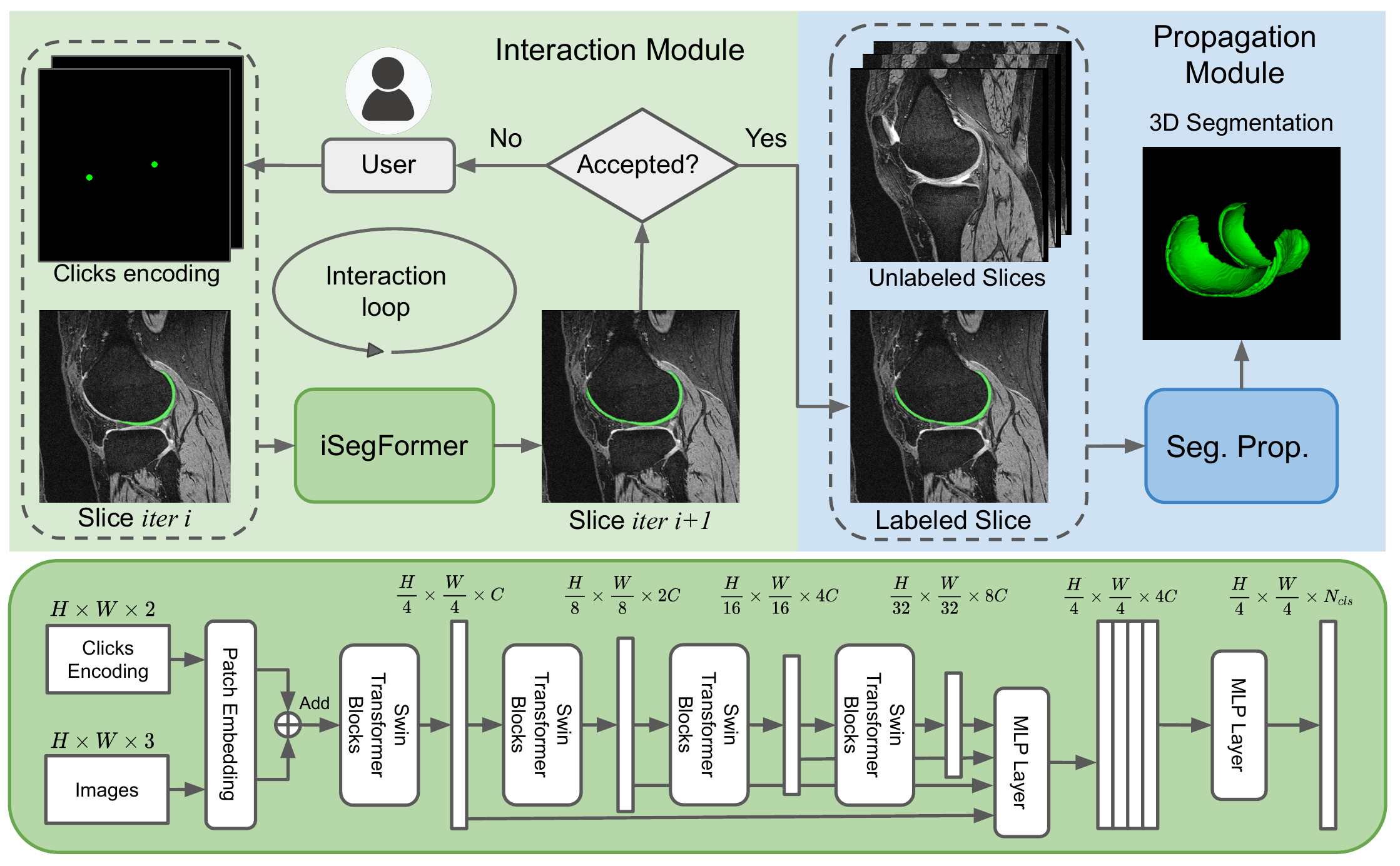}
  \caption{Illustration of {\iSegFormer} for interactive 3D medical image segmentation. Initially, the user selects and interactively annotates one slice (i.e., the user clicks on points) to produce a segmentation, followed by propagating the segmented slice to the unsegmented ones. The {\iSegFormer} architecture is shown at the bottom. The output of {\iSegFormer} will be upsampled to the orignal image size.}
  \label{fig:iSegFormer}
\end{figure}

The proposed {\iSegFormer} is a Transformer-based interactive 2D image segmentation approach that combines a Swin Transformer with a lightweight MLP decoder. 
As shown in Fig.~\ref{fig:iSegFormer}, it can be easily extended to a 3D interactive segmentation approach by combining it with a segmentation propagation module (i.e., STCN~\cite{cheng2021rethinking}). This 3D interactive segmentation approach consists of an {\iSegFormer} for obtaining 2D segmentation from user interactions and a segmentation propagation module that propagates segmented slices to unsegmented ones, resulting in a 3D segmentation.
If the propagated segmentation results are not desired, the user can refine them with further interactions and start a new round of propagation if necessary.
\subsection{Network Architecture of {\iSegFormer}}
\label{sec:architecture_of_isegformer}

The network architecture of {\iSegFormer} is shown in Fig.~\ref{fig:iSegFormer} (bottom). It uses a Swin Transformer as the segmentation backbone and two light-weight MLP layers as the decoder to produce segmentation.
Specifically, there are four Swin Transformer blocks for hierarchical self-attention and a simple MLP decoder that first aggregates both local and global attention and then produces the segmentation as the output.
The input of {\iSegFormer} is the concatenation of image and clicks encoding map (introduced in Sec.~\ref{sec:clicks_encoding_and_slimulation}). 
Since we want to make use of existing pretrained Swin Transformer models on ImageNet-21k~\cite{deng2009imagenet}, we do not change the number of input channels of the Swin Transformer blocks.
To achieve this, we use element-wise addition instead of concatenation for merging image features and clicks encoding features after the patch embedding layers, which are linear projection layers that produce patch embeddings for self-attention. Note that there are two separate patch embedding layers in {\iSegFormer} (one for the input image and the other for the clicks encoding map), though Fig.~\ref{fig:iSegFormer} only shows one for brevity.
The clicks embedding is essential for extending a segmentation model to an interactive segmentation model as it transforms user's interactions from clicks to feature maps that can be fed into the network.
For medical images which typically only have one gray channel, we simply replicate the gray channel to RGB format.

\subsection{Clicks Encoding and Simulation}
\label{sec:clicks_encoding_and_slimulation}

We use clicks as the interaction mode due to their simplicity. 
Clicks can be either positive or negative: positive clicks indicate that particular points should be included in the segmentation, and negative clicks indicate that particular points should not be included in the segmentation. 
We encode positive and negative clicks from coordinates to a 2-channel feature map with the same spatial size with the input image, following the strategy used in ~\cite{benenson2019large}.
The clicks encoding map will be fed into the network along with the input image, as shown in Fig.~\ref{fig:iSegFormer} (bottom).
During training and inference, we automatically simulate clicks based on the ground truth and current predicted segmentation for fast training and evaluation. A positive click is generated in the center of the false negative region in the predicted segmentation, and a negative click is generated in the center of the false positive region in  the predicted segmentation. During training, we add random perturbations for the simulated clicks to increase robustness, as adopted in ~\cite{sofiiuk2021reviving}. During inference, we remove the randomness for determinstic evaluation.
Note that clicks simulation requires the ground truth, and simulated clicks may be different from clicks generated by human evaluation.
Therefore, we present in the supplementary materials some qualitative results obtained by human evaluation.

\subsection{Training and Inference Details}
\label{sec:training_and_inference_details}

For fair comparison with RITM~\cite{sofiiuk2021reviving}, we adopt most of the hyper-parameters used in RITM for training and inference.
The {\iSegFormer} models are trained in a class-agnostic binary segmentation task with the normalized focal loss function (NFL)~\cite{sofiiuk2019adaptis}. 
We randomly crop the image to the size of $320\times480$ for training. 
We adopt the same data augmentation techniques with RITM~\cite{sofiiuk2021reviving} including random scaling and resizing. 
We implement {\iSegFormer} using Pytorch with Adam optimizer. 
All experiments are conducted on a NVIDIA A6000 GPU.
All models are trained 55 epochs with batch size as 32 (except the SegFormer and HRFormer models in Fig.~\ref{tab:comparison_with_other_transformer_backbones}).
More details please refer to our codebase.



\subsection{Extending to Interactive 3D Image Segmentation}
\label{sec:extending_to_interactive_3D_image_segmentation}
{\iSegFormer} can be easily extended to a 3D interactive segmentation approach by combining it with a segmentation propagation module (i.e., STCN~\cite{cheng2021rethinking}).
Since this is not our main contribution, we introduce the details in the appendix.


\section{Experiments}
\vspace{-0.1cm}
\subsubsection{Datasets} The OAI-ZIB~\cite{ambellan2019automated} dataset consists of 507 3D MR images with segmentations for femur, tibia, tibial cartilage, and femoral cartilage.
In this work, we only consider cartilage segmentation. Each 3D image contains 160 slices of size of 384$\times$384. 
We split the dataset randomly into 407 images for training, 50 images for validation, and 50 images for testing.
Since we are interested in the problem setting where the segmentations for the 3D images are limited, we only use three segmented slices of each image in the training and validation sets for developing {\iSegFormer}, resulting in 1521 training slices, 150 validation slices, and 150 testing slices. The three slices are selected at a fixed interval (ie., slice 40, 80, and 120).
We also use 9 other public datasets in our cross-domain evaluation experiments. Please refer to Sec.~\ref{sec:results_on_interactive_2d_image_segmentation} for details.
\subsubsection{Evaluation Metrics}
We use Number of Clicks (NoC) to measure the number of clicks required to achieve a predefined Intersection over Union (IoU) between predicted and ground truth segmentations. For example, NoC@85\% measures the number of clicks required to obtain 85\% IoU. We use an automatic evaluation procedure to simulate clicks during inference and report the quantitative results, following the practices used in~\cite{sofiiuk2021reviving}.
We also perform a human evaluation for a qualitative study.
For measuring the 3D segmentation results, we use the Dice Similarity Coefficient (DSC), sensitivity (SEN), and the positive predictive value (PPV).

\subsection{Results of Interactive 2D Image Segmentation}
\label{sec:results_on_interactive_2d_image_segmentation}
We compare {\iSegFormer} with RITM~\cite{sofiiuk2021reviving}, the state-of-the-art CNN-based approach for interactive 2D femoral and tibial cartilage segmentation.
Both RITM and {\iSegFormer} are implemented on two segmentation backbones.
For RITM, the backbones are UNet~\cite{ronneberger2015unet} and HRNet32~\cite{wang2020hrnet}.
For {\iSegFormer}, the backbones are Swin Transformer's base and large models.
For fair comparison, all the models are trained on the OAI-ZIB training set under the same training settings.

\begin{table}
\footnotesize
\centering
\begin{tabular}{l c c c c c c c c}
    \toprule
    Model & Mem (M) & SPC (ms) & NoC@80\% & NoC@85\% & NoC@90\% & $\geq$20@85 & $\geq$20@90\\
    \midrule
    RITM-UNet & 2680 & 56 & 9.74 (8.76) & 15.28 (7.39) & 17.79 (3.54) & 102 & 144 \\
    RITM-HR32 & 2763 & 82 & 8.47 (8.12) & 14.48 (7.82) & 18.85 (2.47) & 94 & 138 \\
    \midrule
    Ours-Swin-B & 2797 & 64 & 7.25 (7.74) & \textbf{11.67} (8.19) & \textbf{17.03} (6.05) & \textbf{68} & \textbf{115} \\
    Ours-Swin-L & 3755 & 89 & \textbf{6.91} (7.59) & 11.77 (8.26) & 17.57 (5.45) & 70 & 123\\
    \bottomrule
\end{tabular}
\caption{Evaluation on the OAI-ZIB test set for femoral and tibial cartilage segmentation. ``Mem'' denotes the GPU memory consumption for inference. ``SPC'' represents second per click. ``$\geq$20@85'' measures the number of difficult cases that require more than 20 clicks to achieve 85\% IoU. We report mean and standard deviation for the NoC metrics.}
\label{tab:interaction_to_mask}
\end{table}

Tab.~\ref{tab:interaction_to_mask} reports the comparison results for tibial and femoral cartilage segmentation on the 150 slices of the OAI-ZIB testing set.
The results show that {\iSegFormer} outperforms its CNN counterparts with very competitive speed and GPU memory consumption, demonstrating the effectiveness and efficiency of {\iSegFormer} for interactive segmentation.

\subsubsection{Comparison with Other Transformer Backbones}
\label{sec:memory_comparision}
To further demonstrate the efficiency of {\iSegFormer}, we also implemented {\iSegFormer} using two recently proposed Transformer backbones for segmentation: HRFormer~\cite{yuan2021hrformer} and SegFormer~\cite{xie2021segformer}. 
As shown in Tab~\ref{tab:comparison_with_other_transformer_backbones}, our proposed Swin Transformer-based segmentation backbone is much more memory-efficient than the other Transformer-based backbones.

\subsubsection{Cross-Domain Evaluation}
We have shown that {\iSegFormer} outperforms CNNs when trained with only 1,221 labeled 2D slices (labeling such a dataset amounts to labeling 8 3D images with 160 slices).
However, in many applications no segmented slices are available, for example, when studying new medical image datasets.
Therefore, it is important to generalize the trained interactive segmentation models to unseen objects or objects in different domains.
In this cross-domain evaluation, we train {\iSegFormer} and RITM models on the COCO+LVIS~\cite{lin2014microsoft} dataset, which contains millions of high-quality labels for natural images. Then we test the model on 5 natural image datasets (GrabCut, Berkeley, DAVIS, PascalVOC, and SBD) and 3 medical image datasets (ssTEM, BraTS, and OAI-ZIB).

\begin{figure}
  \centering
  \includegraphics[width=8.5cm, height=4.3cm]{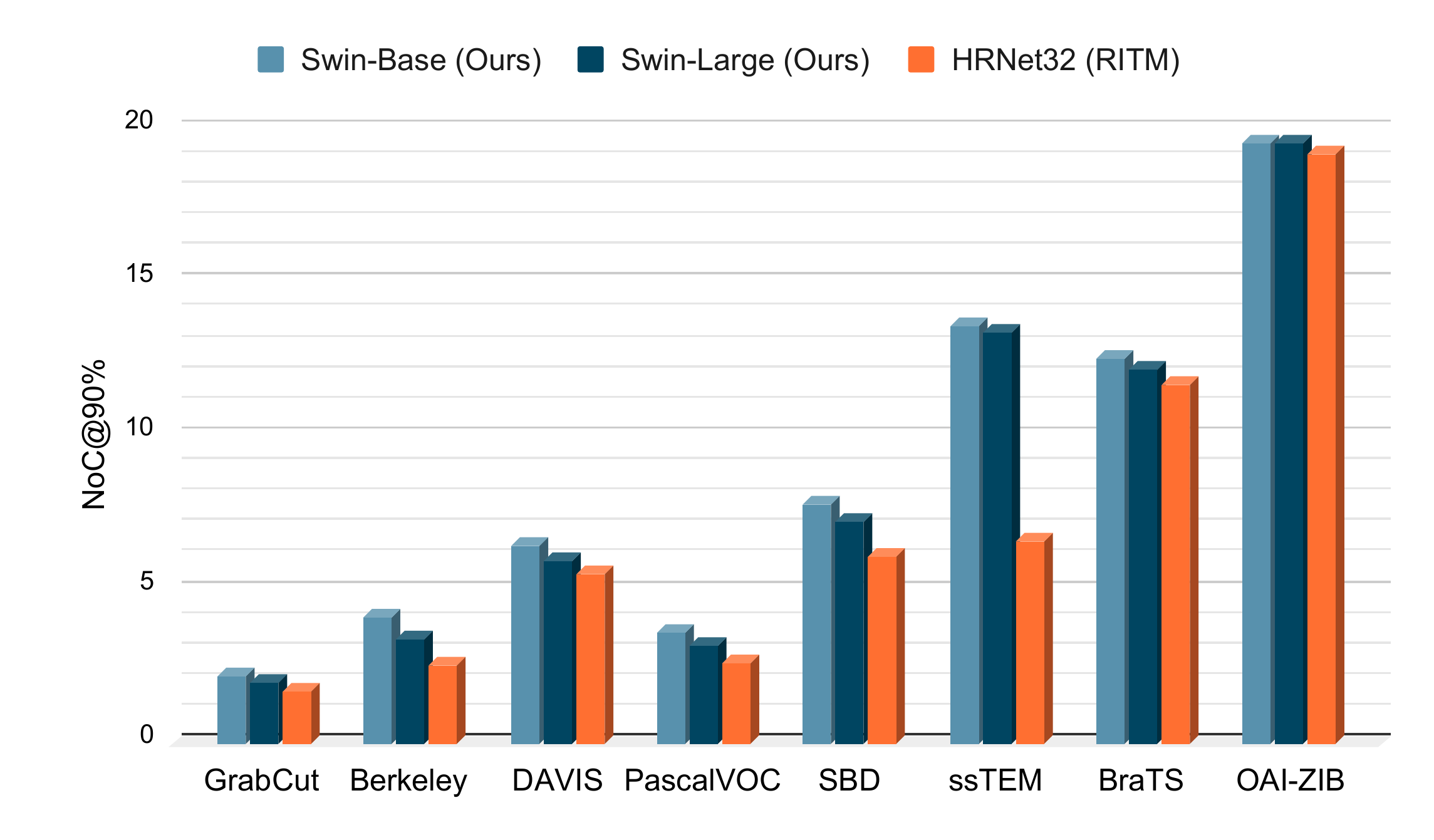}
  \caption{Cross-domain evaluation results. Models are trained on the COCO+LVIS dataset and tested on 5 natural datasets (GrabCut, Berkeley, DAVIS, PascalVOC, and SBD) and 3 medical image datasets (ssTEM, BraTS, and OAI-ZIB).}
  \label{fig:extensitve_evaluation}
\end{figure}

The results are shown in Fig.~\ref{fig:extensitve_evaluation}.
Although there is still a significant performance gap between in-domain and out-of-domain evaluations, both CNN and Transformer models generalize reasonably well to medical image datasets.
Note that our models do not outperform the CNN counterpart in this experiment.
We argue that HRNet is the best performing model in RITM with well-tuned hyper parameters, while we adopt most of their hyper parameters for Transformers and spend little effort in tuning them.

\subsection{Results on Segmentation Propagation}
Given the interactively segmented 2D slices obtained by {\iSegFormer}, we now interested in 3D segmentation via a segmentation propagation model released by STCN~\cite{cheng2021rethinking}. 
The results in Tab.~\ref{tab:results_on_segmentation_propagation} show that with more segmented slices, the propagation results get better.
With only 5 segmented slices, it achieves a Dice score of 82.2\% for femoral cartilage segmentation. 
This is a very promising result considering that the segmentation propagation model was not trained on the medical images.
We hope this preliminary experiment would attract more research effort in transferring knowledge from video domain to the medical imaging domain.

\begin{table}
\footnotesize
\parbox{.45\linewidth} {
\centering
\begin{tabular}{c c c c c}
    \toprule
    \#Slices & DSC (\%) & SEN (\%) & PPV (\%) & IoU (\%) \\
    \midrule
    1  & 55.1  & 61.7 & 55.3 & 38.0 \\
    3  & 78.7  & 85.8 & 73.1 & 64.9 \\
    5  & 82.2  & 87.1 & 77.9 & 70.6 \\
    10 & 85.1  & 88.1 & 88.3 & 74.1 \\
    \bottomrule
\end{tabular}
\caption{Femoral cartilage segmentation using 1, 3, 5, or 10 segmented slices for propagation. The propagation model is trained on video.}
\label{tab:results_on_segmentation_propagation}
}
\hfill
\parbox{.45\linewidth} {
\centering
\begin{tabular}{l c c c}
    \toprule
    Backbone & Params (M) & Mem & SPC \\
    \midrule
    HRNet32 & 41 & 2763M & 82ms \\
    SegFormer-B5 & 28 & $>$5G & $>$0.2s \\
    HRFormer-B  & 50 & $>$5G & $>$0.2s \\
    Swin-B    & 88 & 2797M & 64ms \\
    Swin-L    & 197 & 3755M & 89ms \\
    \bottomrule
\end{tabular}
\caption{Memory and speed comparison between different segmentation backbones.}
\label{tab:comparison_with_other_transformer_backbones}
}
\end{table}

\subsection{Ablation Study}
We demonstrated in Sec.~\ref{sec:results_on_interactive_2d_image_segmentation} that {\iSegFormer} performed better than its CNN counterparts.
Other than the architecture difference, the biggest difference comes from the pretraining settings.
In Sec.~\ref{sec:results_on_interactive_2d_image_segmentation}, our {\iSegFormer} models are pretrained on ImageNet-21k, while the CNN models have two pretraining steps: first pretrained on ImageNet-21k and then finetuned on the COCO+LVIS dataset.
In this study, we adopt different pretrain settings for a more fair comparison between Transformer and CNN models.
Note that the pre-training task can be either classification (Cls) or interactive segmentation (iSeg).
As shown in Tab.~\ref{tab:ablation_study}, pretraining on Image21k is essential for the success of {\iSegFormer}.
More details are included in the supplementary materials.
\begin{table}
\footnotesize
\centering
\begin{tabular}{c c c c c c c c c}
    \toprule
    \multirow{2}{*}{\parbox{1.5cm}{\centering Pretrain Dataset}} & 
    \multirow{2}{*}{\parbox{1.0cm}{\centering Pretrain Task}} & 
    \multirow{2}{*}{\parbox{1.0cm}{\centering Fine Tune}} & 
    \multicolumn{2}{c}{Swin-B} & 
    \multicolumn{2}{c}{Swin-L} & 
    \multicolumn{2}{c}{HRNet32} \\
    & & & NoC@80 & NoC@85 & NoC@80 & NoC@85 & NoC@80 & NoC@85 \\
    \midrule
    N/A & N/A & \cmark & 19.69 & 19.99 & 18.85 & 19.91 & 15.47 & 18.99 \\
    ImageNet-21K & Cls & \cmark & 7.25 & 11.67 & 6.91 & 11.77 & 17.19 & 19.61 \\
    COCO+LVIS    & iSeg & \xmark & 15.48 & 17.19 & 15.09 & 17.45 & 14.58 & 16.93 \\
    COCO+LVIS    & iSeg & \cmark & 12.11 & 15.73 & 9.00 & 13.29 & 8.47 & 14.48 \\
    OAI (w/o GT) & iSeg & \xmark & 18.49 & 19.65 & 18.89 & 19.72 & 16.35 & 18.48 \\
    OAI (w/o GT) & iSeg & \cmark & 12.67 & 15.87 & 13.01 & 16.41 & 7.93 & 12.81 \\
    \bottomrule
\end{tabular}
\caption{Ablation study on pretraining strategies. The pretraining task can be either classification (Cls) or interactive segmentation (iSeg).}
\label{tab:ablation_study}
\end{table}



\section{Conclusion}
We proposed {\iSegFormer}, a memory-efficient Transformer that combined a Swin Transformer with a lightweight multilayer perceptron (MLP) decoder. 
{\iSegFormer} outperformed its CNN counterparts on the interactive 2D knee cartilage segmentation while achieving comparable computational efficiency with CNNs. 
We further extended {\iSegFormer} for interactive 3D knee cartilage segmentation by combining it with a pre-existing segmentation propagation model trained on videos, achieving promissing results even without finetuning the segmentation propagation model.

\section*{Acknowledgement}
Research reported in this publication was supported by the National Institutes of Health (NIH) under award number NIH 1R01AR072013. The content is solely the responsibility of the authors and does not necessarily represent the official views of the NIH.
Data related to knee osteoarthritis used in the preparation of this manuscript were obtained from the controlled access datasets distributed from the Osteoarthritis Initiative (OAI), a data repository housed within the NIMH Data Archive (NDA). OAI is a collaborative informatics system created by the National Institute of Mental Health and the National Institute of Arthritis, Musculoskeletal and Skin Diseases (NIAMS) to provide a worldwide resource to quicken the pace of biomarker identification, scientific investigation and OA drug development.  Dataset identifier(s): NIMH Data Archive Collection ID: 2343.

{\small
\bibliographystyle{ieeetr}
\bibliography{iSegFormer}
}

\end{document}


\title{Supplementary Material}

\section{Datasets}
This section shows the details of the 9 datasets used in the cross-domain evaluation experiment.
\textbf{GrabCut~\cite{rother2004grabcut}} contains 50 images with 50 instance masks. It is widely used for the evaluation of interactive segmentation methods; \textbf{Berkeley~\cite{martin2001berkeley}} contains 96 images with 100 instance masks. It shares a small portion of images with the GrabCut dataset; \textbf{DAVIS~\cite{perazzi2016davis}} contains 50 videos with high quality segmentation masks. Instead of using the entire dataset, we only extract the same 10\% of frames that were used in~\cite{jang2019interactive} for evaluation. \textbf{SBD~\cite{hariharan2011semantic}} contains 6,671 instance-level masks for 2,820 images. This dataset shares the categories with the PASCAL dataset. We use its training set for training and its validation set for evaluation;
\textbf{COCO~\cite{lin2014microsoft}} contains a total of 1.2M instance masks on 118k training images with 80 object categories. We combine this dataset with the LVIS dataset as a training set;
\textbf{LVIS~\cite{gupta2019lvis}} shares its images with the COCO dataset but has the highest annotation quality among all the reported datasets on more than a thousand object categories;
\textbf{PASCAL~\cite{everingham2010pascal}}: we use the validation set for testing, which contains 1,449 images with 3,427 instances;
\textbf{ssTEM~\cite{gerhard2013segmented}} consists of two image stacks where each contains 20 sections from serial section Transmission Electron Microscopy (ssTEM) of the drosophila melanogaster third instar larva ventral nerve cord. We only use the first stack and the mitochondria mask for evaluation;
\textbf{BraTS~\cite{baid2021rsna}} contains 369 training volumes with multi-label annotation masks. We only consider the tumor core; we extract one slice from each volume where the tumor area is largest. This results in 369 slices with binary masks for evaluation.

\section{Method}
\subsection{Differences Between iSegFormer and SegFormer}
Though the names are similar, iSegFormer significantly differs from SegFormer~\cite{xie2021segformer} in the following aspects:1) iSegFormer is an interactive segmentation model while SegFormer is not; 2) iSegFormer adopts the Swin Transformer as the encoder, and therefore is much more efficient than SegFormer.

\subsection{Segmentation Propagation}

In this work, we integrate STCN~\cite{cheng2021rethinking}, an existing state-of-the-art segmentation mask propagation module, into {\iSegFormer} to extend {\iSegFormer} to interactive 3D medical image segmentation. We also adopt an existing bidirectional propagation strategy for segmentation propagation, similar to~\cite{cheng2021modular,oh2019fast}. Hence, while our methodological contribution for segmentation propagation is small, we do show that this combination of techniques works surprisingly well on medical images (though the propagation model was trained on videos), demonstrating the potential for future research in transferring learning for segmentation propagation.

\section{Ablation Study Details}

In the ablation study section of the main paper, we ablate the pretraining settings for {\iSegFormer} and CNN models. That is, we train models use different training sets and tasks. Once the model is pretrained, it will be fine tuned on the OAI-ZIB training set. 
For the COCO+LVIS dataset, we use its high-quality ground truth for training the interactive segmentation model. Since the OAI dataset has no ground truth other than its subset OAI-ZIB dataset, we use SuperPixel~\cite{achanta2012slic} for generating pseudo ground truth for all the OAI dataset. 
We observe that fine tuning helps for all the models.
we also observe that the best model is pretrained on ImageNet-21k dataset, not on the OAI dataset or the COCO+LVIS dataset. 
This may be caused by the inaccurate psuedo labels in the OAI dataset or the out-of-domain bias introduced by the COCO+LVIS dataset.

{\small
\bibliographystyle{ieeetr}
\bibliography{iSegFormer}
}